\documentclass[letterpaper, 10 pt, journal, twoside]{IEEEtran}  %

\IEEEoverridecommandlockouts                              

\usepackage{graphicx} 
\usepackage{amsmath} 
\usepackage{amssymb}  
\usepackage[usenames, dvipsnames]{color}

\usepackage{lipsum}
\usepackage{color}
\usepackage{cite}

\usepackage{diagbox}
\usepackage{tabu}
\usepackage{balance}  

\usepackage[ruled,linesnumbered]{algorithm2e}
\usepackage{algpseudocode}
\usepackage{epsfig}
\usepackage{multirow}

\usepackage{enumitem}
\usepackage{setspace}
\usepackage{tikz}
\usepackage{url}
\usepackage{svg}
\usepackage{makecell}
\usepackage{booktabs}
\usepackage[T1]{fontenc}

\title{
Class-Aware Cartilage Segmentation for Autonomous US-CT Registration in Robotic Intercostal Ultrasound Imaging
}

\author{Zhongliang Jiang*, Yunfeng Kang*, Yuan Bi, Xuesong Li, Chenyang Li, and Nassir Navab, \textit{Fellow, IEEE} 
\thanks{$^{*}$ Authors with equal contributions.}
\thanks{Z. Jiang, Y., Kang, Y. Bi, X. Li, C. Li and N. Navab are with the Chair for Computer Aided Medical Procedures and Augmented Reality (CAMP), Technical University of Munich (TUM), 85748 Garching, Germany. {\tt\footnotesize{(zl.jiang@tum.de)}}
        }%
\thanks{This work involved human subjects in its research. Approval of all ethical and experimental procedures and protocols was granted by Institutional Review Board, No. 2022-87-S-KK, Declaration of Helsinki.}
}

\begin{document}

\maketitle


\begin{abstract}
Ultrasound imaging has been widely used in clinical examinations owing to the advantages of being portable, real-time, and radiation-free. Considering the potential of extensive deployment of autonomous examination systems in hospitals, robotic US imaging has attracted increased attention. However, due to the inter-patient variations, it is still challenging to have an optimal path for each patient, particularly for thoracic applications with limited acoustic windows, e.g., intercostal liver imaging. To address this problem, a class-aware cartilage bone segmentation network with geometry-constraint post-processing is presented to capture patient-specific rib skeletons. Then, a dense skeleton graph-based non-rigid registration is presented to map the intercostal scanning path from a generic template to individual patients. By explicitly considering the high-acoustic impedance bone structures, the transferred scanning path can be precisely located in the intercostal space, enhancing the visibility of internal organs by reducing the acoustic shadow. To evaluate the proposed approach, 
the final path mapping performance is validated on five distinct CTs and two volunteer US data, resulting in ten pairs of CT-US combinations. Results demonstrate that the proposed graph-based registration method can robustly and precisely map the path from CT template to individual patients (Euclidean error: $2.21\pm1.11~mm$). 
\end{abstract}

\def\abstractname{Note to Practitioners}
\begin{abstract}
The precise mapping of trajectories has been a bottleneck in developing autonomous intercostal intervention within limited acoustic space. Existing methods, based on external features such as the skin surface or passive markers, fail to capture the acoustic properties of local tissues, leading to significant shadowing when ribs are involved. The proposed method begins by utilizing distinctive anatomical features to extract cartilage bones and stiff ribs through a class-aware segmentation network. To ensure the segmentation accuracy of the shape of the anatomy of interest, a VAE-based boundary-constraint post-processing in manifold space is developed. Subsequently, a dense skeleton graph-based registration is developed to explicitly consider the subcutaneous bone structure, allowing for the precise mapping of intercostal paths from generic templates to individual patients. Results from ten randomly paired CT and US datasets show that the proposed method accurately maps the intercostal path from the template to individual patients, significantly improving accuracy and robustness over previous methods. We believe that the proposed method can further pave the way for autonomous robotic US imaging. 
\end{abstract}


\begin{IEEEkeywords}
US bone segmentation, intercostal ultrasound scaning, ultrasound segmentation, robotic ultrasound 
\end{IEEEkeywords}



\bstctlcite{IEEEexample:BSTcontrol}

\section{Introduction}

\begin{figure}[ht!]
\centering
\includegraphics[width=0.45\textwidth]{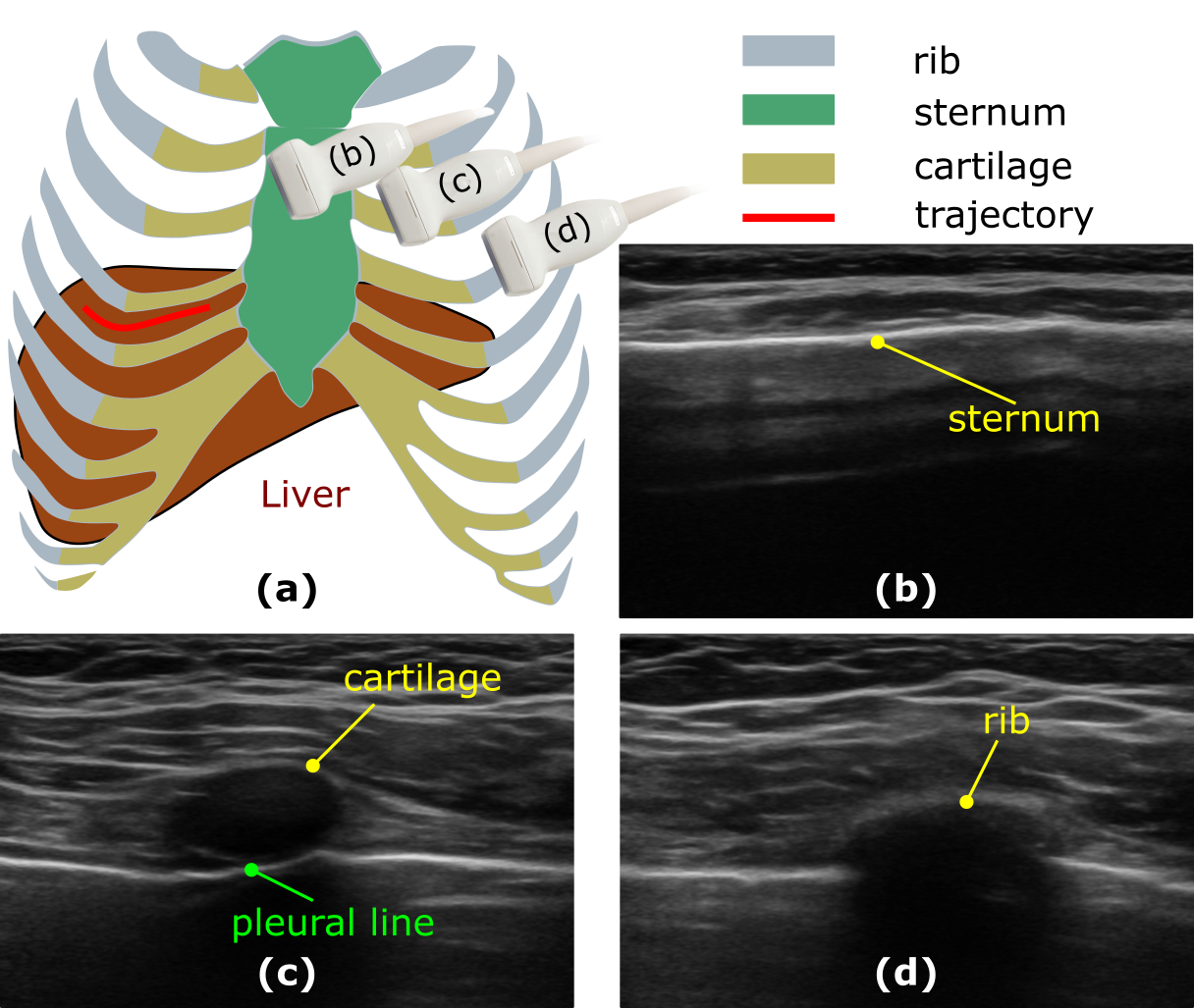}
\caption{(a) Illustration of US liver scan from intercostal space and three types of thorax bones: sternum, rib and costal cartilage. (b), (c) and (d) are the representative US images acquired on the sternum, rib and cartilage, respectively. They have distinct anatomical features on US images.
}
\label{Fig_liver_ablation}
\end{figure}

\par
\IEEEPARstart{M}{edical} ultrasound (US) has been widely used in the preliminary healthcare industry due to its advantages of non-ionizing radiation, real-time capability, and accessibility. Besides the examination of internal organs, US also plays a crucial role in image-guided therapies such as liver ablation~\cite{tsang2021high, kim2014sonographic}. A representative US-guided radiofrequency ablation (RFA) procedure through intercostal space is depicted in Fig.~\ref{Fig_liver_ablation}. Since the bone has much larger acoustic impedance than soft tissues, the US probe should be precisely positioned in the intercostal space to provide a good imaging window. In addition, to avoid penetrating intercostal vessels in liver ablation, electrode or needle should cautiously penetrate through the middle portion of the intercostal space~\cite{kim2015ultrasound}. Due to the fact that hepatic tumors can be adjacent to large vessels or heat-vulnerable organs, the position of the intervention trajectory needs to be very precise. 

\par
To precisely maneuver a US probe, robotic techniques are used frequently owing to its accuracy and repeatability~\cite{jiang2023intelligent}. The comprehensive applications can be found in recent survey articles~\cite{bi2024machine, jiang2023robotic}. In order to develop an autonomous robotic US system (RUSS), Huang~\emph{et al.} computed a multiple-line trajectory based on an external RGBD camera~\cite{huang2018robotic}. To enhance the representation of 3D objects, Tan~\emph{et al.} planned a path on fused surface point clouds captured from multiple cameras~\cite{tan2022fully,tan2022flexible}. However, these methods only consider the outer surface, whereas the planned path cannot guarantee the visibility of the thoracic organs, such as the liver. To address this challenge, Sutedjo~\emph{et al.} computed a scanning path with varying orientations to enhance the coverage level of objects on a phantom with a mimicked rib cage~\cite{sutedjo2022acoustic}. Considering real-world scenarios, G{\"o}bl~\emph{et al.} computed the optimal scanning path for covering liver or heart through intercostal space on tomographic images~\cite{gobl2017acoustic}.

\par
However, it remains to be challenging to accurately transfer the planned path from pre-operative to individual patients. To this end, Hennersperger~\emph{et al.} optimized the registration matrix based on the skin surface point clouds from a live camera and a template~\cite{hennersperger2016towards}. Considering inter-patient variations, Virga~\emph{et al.} used a non-rigid registration approach to further optimize the accuracy of the transferred trajectory~\cite{virga2016automatic}. Specific to the articulated motions of limbs, Jiang~\emph{et al.} applied non-rigid registration to generate patient-specific scanning paths~\cite{jiang2022towards}. These approaches are proven to be robust for their applications (abdominal aorta and limb artery). Nevertheless, their effectiveness on thoracic applications requiring the view through limited intercostal spaces is limited. To address this practical challenge, subcutaneous bone features should be explicitly considered to guarantee the acoustic visibility of internal organs. The bone surface has often been considered as a good reference for such registration because they are not deformed under reasonable pressure applied by US probe~\cite{brossner2023ultrasound}.

\par
To transfer a planned intercostal path from a CT/MRI template to the current setup, Jiang~\emph{et al.} proposed a skeleton graph-based non-rigid registration approach that considers subcutaneous bone structures~\cite{jiang2023skeleton}. This work first utilized the anatomical differences between ribs and cartilage to autonomously select a common region of interest (ROI) across different patients (see Fig.~\ref{Fig_liver_ablation}). In the following work, they further introduced a dense skeleton graph instead of keypoints to reduce the burden of hyper-parameters adjustments while improving the local registration accuracy~\cite{jiang2023thoracic}. However, in these studies, the cartilage bone surface of patients' US acquisitions was manually annotated. An autonomous and robust bone segmentation approach is crucial not only for registration performance but also for validating the feasibility of developing autonomous screening and therapy systems for thoracic applications.

\par
In this study, we present a class-aware cartilage US segmentation Network (CUS-Net) for thoracic US images. To enhance the segmentation quality, a coarse segmentation module and classification module are successively applied. Then, the extracted class activation maps (CAM)~\cite{zhou2016learning} are concatenated with the input images to further do the fine segmentation of cartilage bone images. In the fine segmentation module, both spatial and channel attention mechanisms are applied to enhance segmentation accuracy. To obtain precise anatomy boundary, a geometry-constraint post-processing method is presented based on the variational autoencoder (VAE)~\cite{kingma2013auto}. This is an extension study of our previous idea of dense skeleton graph-based registration work~\cite{jiang2023thoracic} by replacing manual labeling processing using CUS-Net to demonstrate the feasibility of the autonomous intercostal path transferring for RUSS. The main contributions are summarized as follows:

\begin{itemize}
  \item A deep network CUS-Net is proposed to extract the cartilage in coarse-to-fine structure from US images by leveraging classification information.
  
  \item A VAE-based boundary-constraint post-processing in manifold space is presented to enhance the geometry of extracted masks of cartilage US bone. 
  
  \item A dense skeleton graph-based registration is presented to map the scanning path from a generic template to patients by using the autonomously extracted subcutaneous bone features. The method is especially valuable for developing autonomous thoracic scanning programs where acoustic windows (intercostal space) are limited. 
  
\end{itemize}

It is noteworthy that this is the first time that class-aware segmentation and graph-based registration approaches have been combined and jointly evaluated as a complete contribution on unseen volunteers' US and public CT chest volumes\footnote{CT dataset: https://github.com/M3DV/RibSeg} (ten pairs of US-CT combinations). The results demonstrate that the proposed method can significantly outperform the classical ICP, non-rigid ICP, and CPD and Keypoint-based skeleton graph algorithms in terms of Euclidean distance for path transferring error ($2.2\pm1.1~mm$ vs. $13.2\pm9.6~mm$, $5.6\pm2.0~mm$, $6.6\pm3.9~mm$ and $5.6\pm2.5~mm$). The code can be accessed on this webpage\footnote{The code: https://github.com/ge79puv/US\_Cartilage\_Segmentation}.


\par
The rest of this paper is organized as follows. Section II presents related work. The dataset preparation and the implementation details of the CUS-Net are presented in Section III. Section IV describes the details of dense skeleton graph-based non-rigid registration, which was originally presented in our previous conference paper~\cite{jiang2023thoracic}. The experimental results on three volunteers and five CTs are presented in Section V. Finally, the discussion and summary are described in Sections VI and VII, respectively.

\section{Related Work}

\subsection{US Bone Surface Extraction}
\par
Due to the acoustic shadow, poor contrast, speckle noise and inevitable deformation, US image segmentation is a challenging task~\cite{mishra2018ultrasound}. To enhance the quality of bone surfaces (i.e., image contrast), Jiang~\emph{et al.} investigated the impact of probe orientation and suggested that the perpendicular direction of the target's surface will result in better contrast in US bone boundary~\cite{jiang2020automaticTIE, jiang2020automatic}. Hacihaliloglu~\emph{et al.} employed local phase image features as post processing to enhance the appearance of bone surfaces in collected images~\cite{hacihaliloglu2014local, hacihaliloglu2017enhancement}.

\par
To extract the bone boundary from B-mode images, Kowal~\emph{et al.} employed a set of feature-based filters and a grey-level histogram adaptive threshold~\cite{kowal2007automated}. Hacihaliloglu~\emph{et al.} presented a method to automatically determine the contextual parameters of Gabor filters to optimize the local phase methods and they reported that the segmentation performance in terms of surface localization accuracy can be enhanced $35\%$ than the filter with fixed parameters~\cite{hacihaliloglu2011automatic}. In addition, to emphasize the completeness of the bone contour, Wein~\emph{et al.} proposed bone confidence localizer to generate strong responses at possible bone surfaces and low response elsewhere~\cite{wein2015automatic}.

\par
Recently, deep learning has been seen as a promising alternative to the classical feature based approaches. Promising results have been achieved by U-net and its variants on US image segmentation task, such as vessels~\cite{jiang2021autonomous, jiang2022towards}, breast cancer-related lymphedema~\cite{goudarzi2023segmentation} and fetal brain~\cite{venturini2020multi}. Regarding bone segmentation, Salehi~\emph{et al.} applied U-Net structures to generate the probability map and extract the bone boundary using a threshold filter~\cite{salehi2017precise}. To achieve the intensity-invariant performance, Wang~\emph{et al.} used local phase tensor as an guidance to facilitate the bone segmentation on images acquired with different parameters~\cite{wang2020robust}. Besides, Villa~\emph{et al.} intuitively combined the B-mode images with enhanced CPS (confidence map and phase symmetry image) as inputs of a segmentation network~\cite{villa2018fcn}.

\par
Due to the large change in acoustic impedance at the tissue-bone interface, the acoustic shadow is often generated below the interface. Since the shadows are highly related to the bone structure, Alsinan~\emph{et al.} presented a study using a novel generative adversarial network (GAN) architecture to extract the bone shadows and further added the shadow mask as an additional feature to assist the bone surface extraction~\cite{alsinan2020bone}. They reported that introducing an adversarial network improved the generator's performance over the U-net in terms of the Dice coefficient. To preserve bone structure topology, Rahman~\emph{et al.} proposed an orientation-guided graph CNN to ensure the continuity of the segmented bone boundary~\cite{rahman2022orientation}.

\subsection{Coarse-to-Fine Semantic Segmentation}~\label{sec:coarse_to_fine}
\par
In order to enhance the geometry accuracy of segmented objects, the coarse-to-fine framework has been widely used in computer vision tasks by progressively refining the segmentation results. Such methods usually follow the classical detection-then-segmentation strategy. To improve the boundary accuracy, Tang~\emph{et al.} first computed a coarse mask using a segmentation model, and then extracted and refined a series of small image patches along the predicted boundaries using an existing network~\cite{tang2021look}. Similarly, Fu~\emph{et al.} presented a multi-scale recurrent attention network for fine-grained recognition only on category labels, which recursively learns discriminate region and region-based feature representation in a mutually reinforced way~\cite{fu2017look}. 

\par
In the field of medical image analysis, Hu~\emph{et al.} proposed a coarse-to-fine adversarial network architecture to segment extranodal natural killer/T cell lymphoma~\cite{hu2020coarse}. The classical U-Net was first employed to provide the coarse bounding box around the lesion. Then, the refined masks were computed using an end-to-end adversarial network consisting of a U-shape generator and discriminator with the same number of layers as the generator. Specific to US imaging, the segmentation performance suffers from the poor image quality and large variations in the sizes, shapes, and locations of target anatomies. To address these challenges, Wang~\emph{et al.} presented a network with a coarse-to-fine fusion module for accurate US breast tumor segmentation~\cite{wang2021breast}. Instead of the normal skip connections used in U-net, they fuse the latent features in each layer using multiple dilated convolutions providing different perception fields. In addition, Ning~\emph{et al.} proposed SMU-Net to explicitly extract the latent information of background and foreground separately~\cite{ning2021smu}. Then, a fusion module was proposed to recursively fuse the background and foreground feature representatives in each layer. This method achieves superior performance in terms of robustness and accuracy than a few other state-of-the-art methods on US breast datasets.

\subsection{Cross Task Feature Fusion for Semantic Segmentation}
\par
Considering the task to extract the cartilage bone surface for non-rigid registration, a bone classification for individual B-mode images. Since medical image semantic segmentation can be considered as a representative case aiming to extract a pixel-wise classification map, the classification results are considered to be beneficial for enhancing the segmentation accuracy. You~\emph{et al.} proposed a class-aware transformer module to better capture the discriminate regions of object in input images~\cite{you2022class}. To eliminate the need for empirical adjustment of the weight factor for different learning tasks, such as segmentation and classification, Jin~\emph{et al.} proposed entanglement modules to adaptively control the knowledge that can be diffused from one task to another~\cite{jin2021cascade}. The effectiveness of this strategy for boosting multi-task learning had been validated on extensive skin image datasets.

\par
To leverage the intrinsic correlation in segmentation and classification tasks, Xie~\emph{et al.} presented a conjugated network using coarse segmentation to facilitate the classification and then feed the classification result to assist the fine segmentation~\cite{xie2020mutual}. The results demonstrated that such combination can enhance both segmentation and classification accuracy. Zhang~\emph{et al.} designed a feature fusion module to fuse the features obtained by both encoders of segmentation and classification branches~\cite{zhang2022feature}. 

\begin{figure*}[ht!]
\centering
\includegraphics[width=0.75\textwidth]{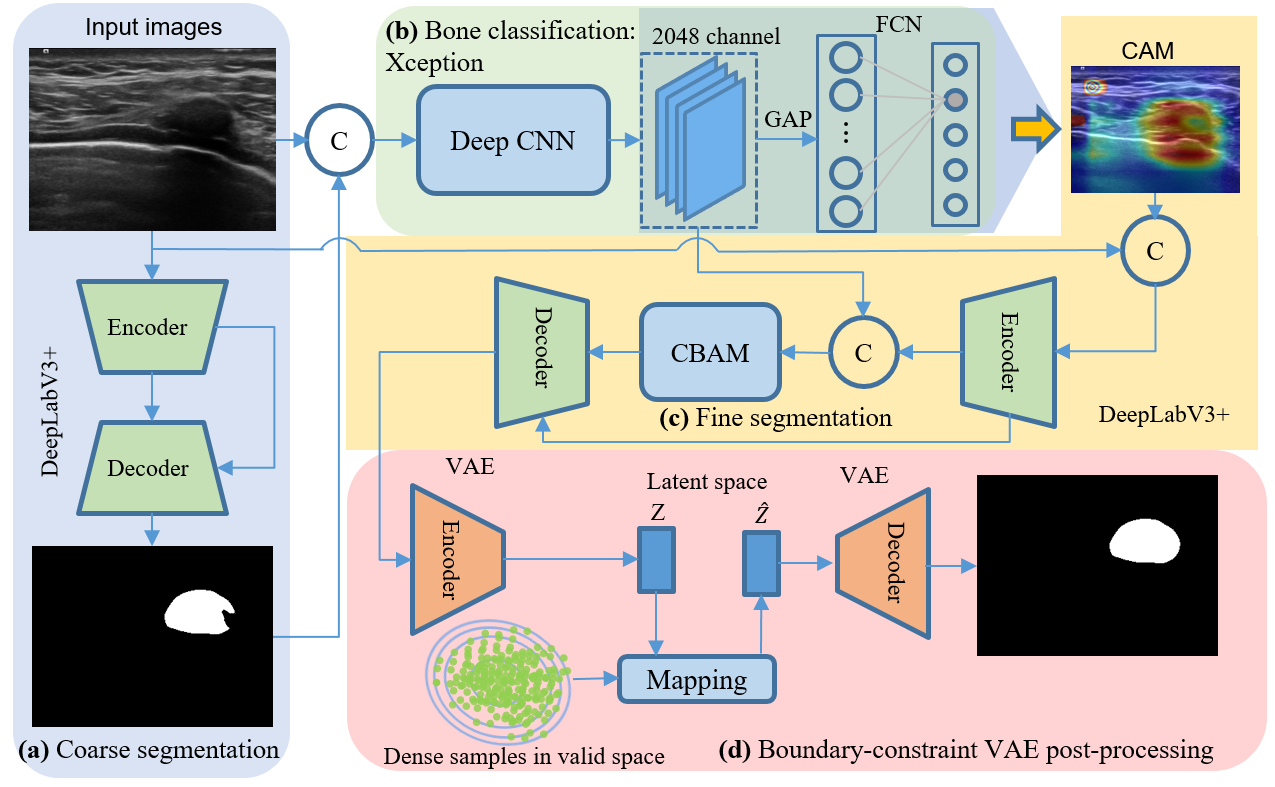}
\caption{The proposed class-aware cartilage bone segmentation architecture. The CUS-Net consists of four distinct modules: coarse segmentation, classification, fine segmentation, and boundary-constraint VAE-based post-processing. First, a coarse segmentation network is employed to generate region proposals for the target anatomy. Subsequently, a classification module is utilized to automatically differentiate between cartilage, rib, and sternum regions. Leveraging the Class Activation Maps (CAM) generated by the classification module, a fine segmentation process is conducted to improve segmentation accuracy. Finally, a boundary-constrained VAE-based post-processing module is applied to refine the shape accuracy of the cartilage bone, ensuring robust inputs for registration.
}
\label{fig_network}
\end{figure*}

\subsection{US-CT Registration}
\par
To compute the registration matrix between CT and US images, there are two streams of methods: image-based approach~\cite{wein2007simulation} and surface-based approach~\cite{wein2015automatic,ciganovic2018registration}. The former directly optimizes the registration based on various image similarity terms, such as LC$^2$~\cite{wein2007simulation}. Lei~\emph{et al.} registered intraoperative 2D US images to 3D CT for needle intervention~\cite{lei2023robotic}. The image-based approaches do not rely one precisely segmentation, but the results often suffer from US imaging noise~\cite{hacihaliloglu2014local}.

\par
In contrast, surface-based approaches~\cite{wein2015automatic,ciganovic2018registration} are built upon precise segmentation. Based on the extracted surface point clouds from both source and target spaces, the classical ICP algorithm~\cite{besl1992method} can be used to compute the transformation matrix. Considering the point clouds may only be partially observed, Zhang~\emph{et al.} incorporate the partially reliable normal vectors, formulating the registration problem as a maximum likelihood estimation problem~\cite{zhang2021reliable}. Experiments on a femur head demonstrated that the method could robustly and accurately optimize the rigid matrix. It is worth noting that the performance of this method may degrade if the surface of the target anatomy is relatively flat.

\section{Class-Aware Cartilage US Segmentation}
\par
The precise US bone surface segmentation is the crucial part of this study. Considering the distinct feature of cartilage bone (with a visible pleural line beneath), it can be used to assist in selecting the same ROIs from different patients' images for non-rigid registration. To this end, this study proposed a class-aware cartilage bone segmentation network CUS-Net in the coarse-to-fine fashion to simultaneously conduct the segmentation and classification tasks for thoracic US images. The network consists of four modules: (1) coarse segmentation, (2) bone classification, (3) fusion-based segmentation refinement, and (4) boundary-constraint VAE post-processing. The overall network architecture is depicted in Fig.~\ref{fig_network}. The descriptions of each module are given in the following subsections. The detailed implementation is public in this webpage\footnote{Code: https://github.com/ge79puv/US\_Cartilage\_Segmentation}.

\subsection{US Thoracic Bone Dataset}~\label{sec_dataset}
\subsubsection{Hardware Setup}
\par
In this study, all US images were recorded from an ACUSON Juniper US machine (Siemens Healthineers, Germany) using a linear probe 12L3 (Siemens Healthineers, Germany). To access US images, a frame grabber (Epiphan Video, Canada) was used to transfer the real-time image from US machine to the main workstation. The US image acquisition frequency was $30~fps$. To properly visualize the bone structure in B-mode images, a default setting provided by the manufacturer was used in this study: MI: $1.13$, TIS: $0.2$, TIB: $0.2$, DB: $60$ dB. Since the ribs of interest are shallow, the imaging depth was set to $35~mm$. 

\par
To provide precise tracking information for each B-mode image, the probe was attached firmly to the flange of a collaborative robotic arm (LBR iiwa 7 R800, KUKA GmbH, Germany). The robot was controlled via a self-developed robotic operation system (ROS) interface and the robot's status is updated at $100~Hz$. Based on robotic kinematics, the tracking stream of the tool center point (TCP) can be obtained. To precisely stack 2D images into 3D space, both spatial and temporal calibration procedures were carried out as in our previous works~\cite{jiang2023skeleton, jiang2023thoracic}. 

\subsubsection{Data Recording and Preparation}
\par
In order to collect tracked images, we manually maneuver the robotic arm to do multiple-line US scans on the front chest of volunteers. In total, $8721$ thoracic bone images ($2194$, $3200$, and $3327$, respectively) were recorded from three volunteers. Considering the characteristic of different bone images (see Fig.~\ref{Fig_liver_ablation}), we only annotated the surface of the rib and sternum, while the cartilage was annotated as the round region covered by the bone surface and pleural line. All the annotations were carefully carried out in ImFusitionSuite (ImFusion AG, Germany) under the close supervision of a US expert. The CUS-Net was trained on $2194$ images form volunteer 1, while tested on two unseen volunteers (weights: $70~kg$ vs$60~kg$, height: $167~cm$ vs $173~cm$, and BMI: $25.1$ vs $20.0$) to show the effectiveness on different patients. The input images, originally sized at $844 \times 632$ pixels, were resized to $320 \times 240$ pixels.

\subsection{Coarse Segmentation}~\label{sec:sub_coarse_segmentation}
\par
Inspired by the idea of ``look closer to see better"~\cite{fu2017look}, a coarse segmentation network is first used to provide the region proposal of the target anatomies. In this study, the state-of-the-art DeepLabV3+~\cite{chen2018encoder} was chosen due to the superior performance in segmentation tasks. Similar to the classical U-net~\cite{ronneberger2015u}, DeepLabV3+ employed the encoder-decoder structure to preserve the details in the predicted masks, such as sharp object boundaries. The encoder employs the powerful classification network Xception~\cite{chollet2017xception} as the backbone by making a few modifications, such as replacing max pooling with depthwise separable convolution with striding. Then, Atrous Spatial Pyramid Pooling (ASPP) is applied on the output of Xception backbone to explicitly control the resolution and increase the perception field~\cite{chen2018encoder}. After this, a $1\times1$ convolution with $256$ filters is applied to compute the encoder output feature map containing $256$ channels and rich semantic information.

\par
To recover object segmentation details, a simple yet effective decoder module was presented in DeepLabV3+\cite{chen2017deeplab}. The 256-channel output of the encoder is first bi-linearly upsampled by a factor of $4$ and then concatenated with the low-level features with the same spatial resolution. To avoid the potential imbalance between low-level feature and encoder output, an $1\times1$ convolution is applied on the low-level features. A few $3\times3$ convolutions are then used to refine the features, followed by another simple bilinear upsampling by a factor of $4$~\cite{chen2017deeplab}. The detailed implementations used in this study can refer to this code\footnote{https://github.com/YudeWang/deeplabv3plus-pytorch/tree/master}.

\par
To train the segmentation network, the Dice loss is computed between the manually annotated ground truth data $Y$ and the binary segmentation mask $\tilde{Y}$.

\begin{equation} \label{eq_dice_score}
	L_{Dice} = 1- \frac{2 | \tilde{Y}  \cap Y |}{ | \tilde{Y} | + | Y |}
\end{equation}

\par
The coarse segmentation was trained on $2194$ US thoracic bone images of volunteer 1. The ratio between the training, validation, and test is $6:2:2$. Adam optimizer was used in this study. The initial learning rate was $2\times10^{-5}$, and it will be decayed by a factor of $2$ if there is no significant change in the consecutive ten iterations. The coarse segmentation network was trained from scratch for $100$ epochs.



\subsection{Bone Classification}
\par
Considering there are three types of thoracic bones involved in this study, we manually classify the recorded US images into five categories: cartilage, rib, sternum, transition part (i.e., the connection part between cartilage bone and ribs or cartilage bone and sternum), and background (i.e., no bone shown in the image). The aim of explicitly separating the transition part is to ensure the classification network can be more accurately and quickly converged to the right distribution for other classes. The image number of each category of volunteer 1 are summarized as follows: $1042$ cartilage, $280$ rib, $191$ sternum, $574$ transition part, and $107$ background (in total $2194$). The training, validation, and testing data sets are identical to the ones used for coarse segmentation. 


\par
Since both classification and segmentation tasks rely on the effective extraction of object representation from images, the coarse segmentation mask can be used as a region proposal for bone classification. To this end, the binary coarse mask is concatenated to the B-mode image as a two-channel input for the classification network (see Fig.~\ref{fig_network}). Considering the outstanding classification performance of Xception~\cite{chollet2017xception} over a larger image dataset comprising $350$ million images, it was used here for identify the cartilage bone. 
Due to the size of our dataset, a pre-trained model on the PASCAL VOC dataset~\cite{everingham2010pascal} was used as initialization; followed by a fine-tuning process based on thoracic US bone images. 



\par
Regarding the classification task, the cross-entropy loss ($L_{CE}$) is computed as follows:
\begin{equation} \label{eq_CE}
    L_{CE} = -\sum_{c=1}^{M} y_{o, c}~\log(p_{o,c})
\end{equation}
where $y$ is the binary indicator ($0$ or $1$) if class label $c$ is the correct classification for observation $o$, $p$ is the predicted probability of observation $o$ belonging to class $c$. To train the classification network, the Adam optimizer was used. The learning rate was $1\times10^{-4}$ and the batch size was $16$. The pre-trained classification network was further trained for additional $50$ epochs to achieve good performance in this study.

\subsection{Classification-Boosted Fine Cartilage Segmentation}
\par
To explicitly leverage the instinct information between segmentation and classification tasks, we compute the class activation maps (CAM)~\cite{zhou2016learning} using global average pooling (GAP). CAM is a generic localizable deep representation of the implicit attention of CNNs on images. The important and discriminative image regions for classification can be highlighted (see Fig.~\ref{fig_network}). Due to the use of GAP rather than global max pooling~\cite{oquab2015object}, the CAM are encouraged to find the extent of the object instead of one single discriminative part. This makes CAM particularly suitable when there may have multiple bones shown in the same B-mode image. It can be seen from Fig.~\ref{fig_network} that the representative CAM result quite precisely annotates the location of the cartilage bone from the input image. It is worth noting that the CAM are 1-channel images. The transferred color version is only for better visualization.

\par
In order to further refine the bone surface, the class-aware localization map and the original B-mode images are concatenated as a 2-channel image input for the fine segmentation network. Then, DeepLabV3+ is employed for fine segmentation. The brief descriptions of the encoder and decoder are given in Sec.~\ref{sec:coarse_to_fine}. Besides the combination of the 2-channel inputs, the high-level CNN feature representations ($2048$ channels) optimized for the classification are concatenated with the encoder feature map ($256$ channels) in latent space to boost the segmentation performance (see Fig.~\ref{fig_network}). To enhance the boundary accuracy, the effective Convolutional Block Attention Module (CBAM)~\cite{woo2018cbam} is used to force the network to focus more on the important regions based on the attention maps computed in both channel and spatial dimension.

\par
Since the performance of cartilage bone segmentation will significantly affect the performance of the non-rigid registration between US cartilage bone point cloud and template cloud, two fine segmentation models were trained separately. One is tailored only for cartilage bone, while the other is for non-cartilage images. The parameters of both models were initialized the same as the coarse segmentation network. In the fine segmentation process, to encourage the network to pay attention to the anatomical (cartilage) boundary as well, the boundary loss function~\cite{kervadec2021boundary} ($\mathcal{L}_{BD}$) is combined to build the joint loss function ($\mathcal{L}_{FineCar}$)as follows:

\begin{equation} \label{eq_FineCar_loss}
	\begin{split}
		\mathcal{L}_{BD} &= 2 \int_{\Delta S}{D_{G}(q)} dq\\
		\mathcal{L}_{FineCar} &= (1-\alpha)\mathcal{L}_{Dice} + \alpha \mathcal{L}_{BD}\\
	\end{split}
\end{equation}
where $\Delta S$ is the region between the two contours of ground truth $G$ and segmentation mask $S$; $D_{G}$ is the distance map with respect to the boundary of $G$ ($\partial G$), i.e., $D_{G}(q)$ compute the distance between a point $q$ and the nearest point on boundary $\partial G$. Since $\mathcal{L}_{BD}$ is supplementary to $\mathcal{L}_{Dice}$ to enhance the boundary accuracy, a small $\alpha$ is used at the beginning and is gradually increased as the training. Following the rebalance strategy~\cite{kervadec2021boundary}, $\alpha$ was initialized to $0.01$ and increased by $0.01$ after each epoch. The training setting is the same as the coarse segmentation. The other training details are the same as the coarse segmentation.

\subsection{Boundary-Constraint VAE Post-processing}~\label{sec:VAE_bounday}
\par
To explicitly guarantee the anatomical accuracy of segmentation results, a post-processing method~\cite{painchaud2020cardiac} developed based on the rejection sampling approach is adapted here. To this end, a VAE~\cite{kingma2013auto} is first trained to learn the latent representation of the ground truth data without any anatomical aberrations. The VAE encoder can project an input $I_x$ to the latent space, and the decoder will recover the latent vector $\Vec{z}$ back into the input space (reconstructed signal $\hat{I}_x$). Then, it is intuitive that we can enhance the anatomical shape of an implausible $I_x$ by mapping the corresponding latent feature $\vec{z}$ to a near but anatomically valid latent vector $\hat{z}$. The effectiveness of this mapping highly relies on the dimension ($N_f$) of the latent feature vector ($2^{N_f}$). Based on the experiments, $N_f$ was empirically determined to be $5$ in our setup, which can preserve enough textual information for reconstruction for the annotated $1042$ binary masks of cartilage images. To train the VAE~\cite{sohn2015learning}, the loss function ($\mathcal{L}_{VAE}$) consists of the binary cross-entropy and Kullback-Leibler (KL) divergence terms for image reconstruction and regularization, respectively.

\begin{equation} \label{eq_Loss_vae}
\mathcal{L}_{VAE} = -\underbrace{\mathbb{E}_{z\sim q(z|x)} \left[\log p(x|z)\right]}_{\text{reconstruction}} + \underbrace{\mathbb{KL}(q(z|x)||p(z))}_{\text{regularization}} 
\end{equation}
where the reconstruction error is represented by the expected negative log-likelihood of the datapoint, and the regularization error is computed by KL divergence between the encoder’s distribution $q(z|x)$ and prior distribution $p(z)$.

\par
The performance of the mapping from $\vec{z}$ to valid latent vector $\hat{z}$ highly relies on the number of valid samples. To augment the valid latent vectors from a determined complex distribution, rejection sampling~\cite{koller2009probabilistic} is employed. The $\textbf{P}(\vec{z})$ is the distribution of the valid latent vectors. Its probability density function (PDF) $f_p(\vec{z})$ can be obtained using the Kernel density estimation approach (built-in scikit package) on the recorded data. In addition, a Gaussian distribution $\textbf{Q}(\vec{z})$ is fitted based on all valid latent vectors. Its PDF is defined as $f_q(\vec{z})$. Then, there is a constant value $K_{rs}$ that can satisfy the following equation: $f_p(\vec{z}) \leq K_{rs} f_q(\vec{z})$. To densely generate samples in the latent vector space, a random uniform distribution $u \backsim U(0,K_{rs} f_q(\vec{z}_{i}))$ is created. According to the rejection sampling, $z_i$ will be kept if $u<f_p(\vec{z}_i)$; otherwise, it will be rejected. Since the augmented data needs to lie in the valid vector space to maintain the valid shape, the rejection law is redefined as follows:

\begin{equation} \label{eq_rejection}
	u<\mathbb{F}[dec({\vec{z}_i})] f_p(\vec{z}_i)
\end{equation}
where $dec{(\vec{z}_i)}$ is the VAE decoder to reconstruct the segmentation map from latent feature $\vec{z}_i$. $\mathbb{F}(\cdot)$ is the shape-aware function with respect to the reconstructed masks, which returns 1 when the input mask is anatomically plausible and zero otherwise. In our case, the shape will be considered not ideal if the reconstructed masks have holes, or disconnected regions with significantly smaller areas than the target of interest. This sampling process was repeated to generate $110K$ new samples in the manifold of valid space. Then, the latent feature vector $\vec{z}$ of an input segmentation mask can be mapped to a valid sampled vector $\hat{z}$ using $K$-nearest neighbors (KNN) approach. 
\section{Graph-based Skeleton Non-Rigid Registration}
\subsection{Cartilage Point Cloud Generation} \label{sec:Point_Cloud_Generation}
\par
\begin{figure*}[ht!]
\centering
\includegraphics[width=0.95\textwidth]{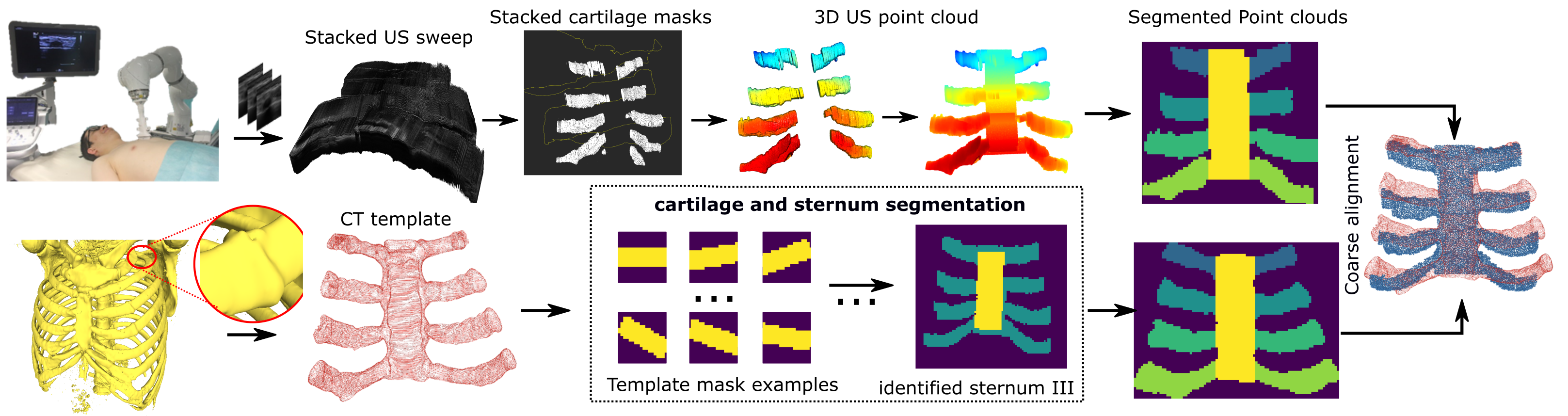}
\caption{Illustration of coarse alignment between CT and US point clouds. The US point cloud is generated based on the autonomous segmented cartilage US images from volunteers and the paired robotic tracking information. The CT point cloud was generated through manual annotation of the CT chest volume. By precisely segmenting the sternum and individual cartilage branches in both the CT template and patient-specific US point clouds, the two point sets can be coarsely aligned by matching the sternum.
}
\label{fig_coarse_alignment}
\end{figure*}

To consider inter-patient variations, a non-rigid registration is required to precisely transfer the scanning path from a generic template to the current setup. It is crucial to use an identical ROI from both CT templates and US images, such as intact organs. Benefiting from the biomarker of cartilage bone on both CT and US images (see Figs.~\ref{Fig_liver_ablation} and~\ref{fig_coarse_alignment}), we can identify and segment the intact cartilage bones from patients. Due to the invariant characteristics of bone, the scanning path planned in the limited intercostal space on the template can be precisely mapped to patients for examination. We extended our dense graph-based cartilage bone registration approach~\cite{jiang2023thoracic} in this study by enabling autonomous segmentation of cartilage.

\subsubsection{Point Clouds Generation from CT Template}
\par
The dense graph-based non-rigid registration is performed on point clouds. The cartilage regions of five patients' CTs were manually obtained by subjectively, 1) applying an intensity threshold-based segmentation in 3D Slicer to extract rib cages (see Fig.~\ref{fig_coarse_alignment}); 2) extracting the ROIs (the cartilage bones of the $2$-nd, $3$-rd, $4$-th, and $5$-th ribs) from CTs based on biomarker on each rib branch in Meshlab; 3) using Poisson disc sampling to generating CT point clouds $\textbf{P}_{ct}$. Due to the limitation of dataset size, a template matching approach is first used to extract the sternum in the projected 2D plane using principal components analysis (PCA). Then, the point clouds of eight cartilage branches are extracted consecutively using the classic K-Nearest-Neighbors (KNN) algorithm (initialized with eight clusters), and flood fill algorithm~\cite{agkland1981edge}. A representative example of segmented $\textbf{P}_{ct}$ is depicted in red in Fig.~\ref{fig_coarse_alignment}. More implementation details can be seen in~\cite{jiang2023thoracic}.

\subsubsection{Point Clouds Generation from US Scans}
\par
Unlike the process for CT point cloud, the US images are simultaneously segmented and classified in this study. Based on the classification and segmentation results, we can stack the cartilage with high accuracy. A representative result with tracking information is visualized in Fig.~\ref{fig_coarse_alignment} (see stacked cartilage masks). Since there are isolated cartilage bone clusters (see the bottom of stacked cartilage masks), which do not belong to the $2$-nd, $3$-rd, $4$-th, and $5$-th ribs, pre-processing is needed to clean the autonomously generated US cartilage point cloud $\textbf{P}_{us}$. To this end, the DBSCAN algorithm~\cite{ester1996density} is used to autonomously identify all clusters based on the density. Based on the performance, the distance to neighbors and minimum points were empirically set to $0.8~cm$ and $16$, respectively. Then, the small clusters are filtered out based on a preset threshold ($3,000$ in this study). Considering the sternum in the CT template is not complete, it only contains the part between $2$-nd, and $5$-th ribs. Therefore, instead of using segmented sternum masks, we directly generate a fake sternum surface by using a rectangle to connect the segmented cartilage ribs. In order to do so, the PCA is applied to unify the acquired $\textbf{P}_{us}$. Based on the centroid of $\textbf{P}_{us}$ and the centroid of each remaining cartilage cluster, the clusters (could be more than eight) can be divided into left and right folders. Then, the sternum between the paired ribs can be generated by connecting the rightmost 2D cartilage plane in the left folder and the leftmost plane in the right folder. Since the path will only be planned in the intercostal space, the thickness of the sternum is less important in this study. A representative process and the final $\textbf{P}_{us}$ can be seen in Fig.~\ref{fig_coarse_alignment}.

\begin{figure*}[htb!]
\centering
\includegraphics[width=0.75\textwidth]{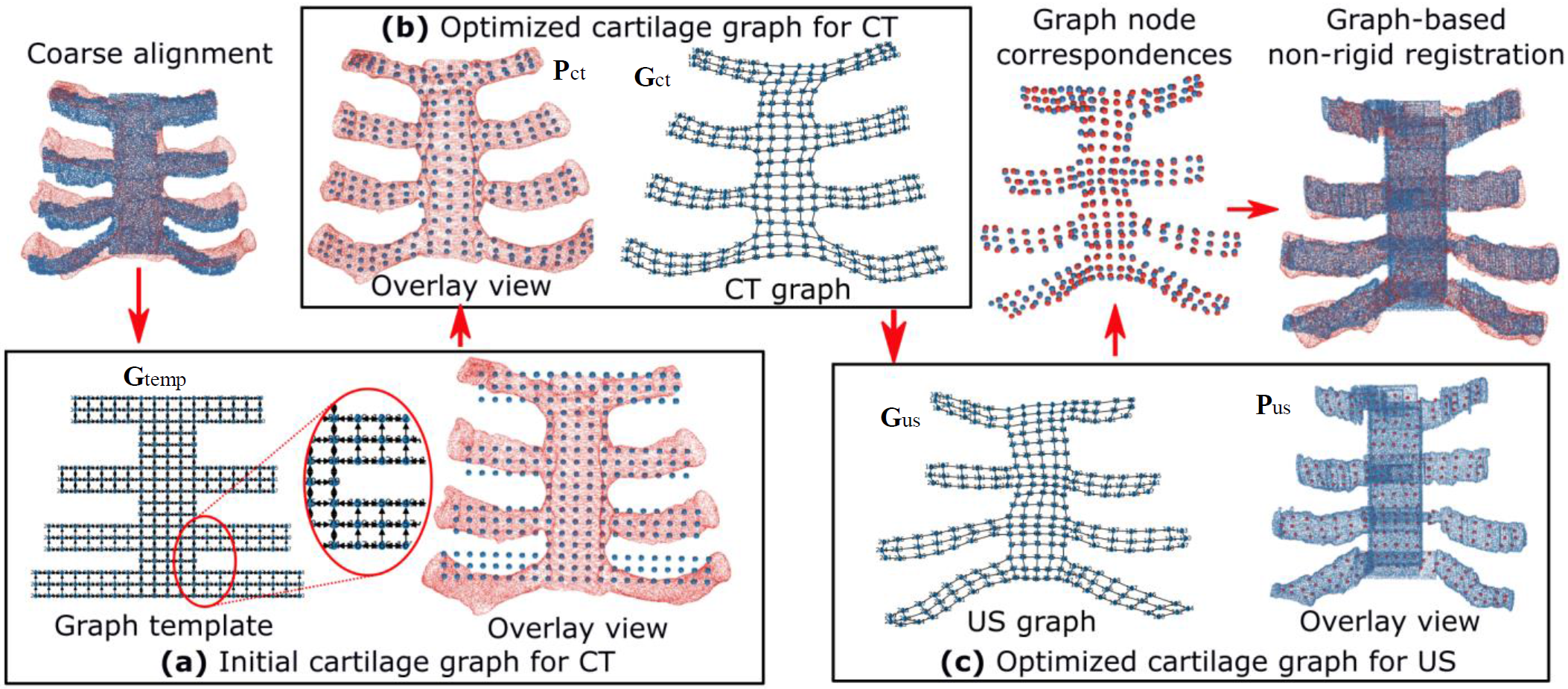}
\caption{The illustration of the fine alignment of CT and US skeleton point clouds using the SOM algorithm based on the geodesic distance. The graph node correspondences can be obtained based on the optimized $\textbf{G}_{ct}$ and $\textbf{G}_{us}$.
}
\label{fig_fine_alignment}
\end{figure*}

\subsection{US-CT Point Clouds Non-Rigid Registration}~\label{sec:graph_registration}
\par
In this section, the graph-based non-rigid registration is elaborated (see Fig.~\ref{fig_fine_alignment}). It is worth noting that the dense skeleton graph-based registration method was originally presented in our previous conference paper~\cite{jiang2023thoracic}. Based on the processed $\textbf{P}_{us}$ and $\textbf{P}_{ct}$, a coarse alignment is carried out at beginning. Then, a modified self-organizing map (SOM) algorithm~\cite{kohonen1990self} is applied twice to obtain two cartilage graphs $\textbf{G}_{us}$ and $\textbf{G}_{ct}$ of $\textbf{P}_{us}$ and $\textbf{P}_{ct}$, respectively. Based on the matched nodes in $\textbf{G}_{us}$ and $\textbf{G}_{ct}$, the planned scanning path in the intercostal space can be transferred from the CT template to the current setup for specific patients.

\subsubsection{SOM-based Graph Node Correspondence Optimization}
\par
Based on the geometry of the given CT templates, a directed template graph $\textbf{G}_{temp}$ is created. Similar to~\cite{jiang2023thoracic}, $\textbf{G}_{temp}$ consists of $245$ evenly distributed nodes. Then, we use $\textbf{G}_{temp}$ as the initial graph for the modified SOM algorithm to characterize the topological structure of a given CT point cloud. Considering the potential misassignment of the nodes among neighbouring cartilage branches, the geodesic distance of directed $\textbf{G}_{temp}$ is used to compute the update rate for moving nodes. The SOM is an unsupervised machine learning method trained using competitive learning. To update $\textbf{G}_{temp}$, the weight vector $\textbf{W}_s$ for each node is calculated between the nodes and a random sample of the input point cloud in terms of a distance metric (here is geodesic distance). The node with the smallest weight is called the best matching unit (BMU). $\textbf{W}_s$ of each node is updated as follows:
\begin{equation}\label{eq_som}
\textbf{W}_{s}(i+1)=\textbf{W}_{s}(i)+\theta_{(BMU, i)}\cdot l_r\cdot \left[\textbf{P}(k)-\textbf{W}_{s}(i)\right]
\end{equation}
where $i$ is the current iteration, $\theta_{(BMU, i)}$ is the updated restriction function computed based on the geodesic distance between BMU and other nodes, $l_r$ is the learning rate, and $\textbf{P}(k)$ is the $k$-th point in the point cloud. 

\par
Since the US point cloud $\textbf{P}_{us}$ has been coarsely aligned with the $\textbf{P}_{ct}$, the optimized CT graph $\textbf{G}_{ct}$ is consecutively used as the initial graph for the SOM algorithm to characterize the topological structure of $\textbf{P}_{us}$. Then, the corresponding nodes in $\textbf{G}_{ct}$ and $\textbf{G}_{us}$ can be paired. The overview of the registration pipeline is depicted in Fig.~\ref{fig_fine_alignment}.

\subsection{Graph-based Non-Rigid Registration for Path Transferring}
\par
Based on the paired node correspondences, the local transformation matrix $^{us}_{ct}\textbf{T}$ mapping $^{ct}\textbf{P}_{g}$ to $^{us}\textbf{P}_{g}$ can be computed by minimizing  Eq.~(\ref{eq_rotatopn_optimization}).

\begin{equation} \label{eq_rotatopn_optimization}
\min_{^{ct}_{us}\textbf{T}} \frac{1}{N_{reg}}\sum_{i=1}^{N_{reg}}{||^{us}_{ct}\textbf{T}~^{ct}\textbf{P}_{g} - {^{us}\textbf{P}_{g}}||^2}
\end{equation}
where $^{ct}\textbf{P}_{g}$ and $^{us}\textbf{P}_{g}$ are the spatial location of paired nodes in $G_{ct}$ and $G_{us}$, respectively. The hyperparameter $N_{reg}$ is empirically set to three based on the experimental performance. A large $N_{reg}$ will reduce the non-rigid property. 
To preserve the anatomy continuity, a weighted transformation approach is employed to transfer the CT point cloud to US space as follows: 
\begin{align}\label{registration_eq}
    ^{ct}\mathbf{P}^{\prime}=\sum_{i=1}^N \frac{d_i}{\sum_{j=1}^N d_j}\left(^{us}_{ct}\textbf{T}_i [^{ct}\mathbf{P}; 1]^{T}\right)
\end{align}
where $d_{i,~\text{or}~j}$ is the Euclidean distance between individual point among $\textbf{P}_{ct}$ and the closet $N$ nodes in $G_{ct}$, $^{ct}\mathbf{P}^{\prime}$ is the transformed CT point in US space.

\par
To map the planned path from the CT space to US space, we need to get enough paired node corresponds. Considering the path is planned in the intercostal spaces, a sphere around each waypoint of the trajectory is created to include enough points in the local area to compute the local transformation matrices. Based on the experimental performance, the sphere radius is empirically set to $20~mm$ in this study. Then, each waypoint of the intercostal scanning path in CT space can be mapped to US space based on the paired point sets from $\textbf{P}_{ct}$ and transferred $^{ct}\textbf{P}^{\prime}$ described in US space.



\section{Results}
\subsection{Bone Classification Performance}
\par
In order to evaluate the performance of the bone classification network, the metrics of accuracy, sensitivity, specificity, and Area Under the Receiver Operating Characteristic (ROC) Curve (AUC) are employed in this study. The quantitative results on $1400$ US bone images from two unseen volunteers are summarized in Table~\ref{tab:classification}. For each volunteer, we selected $50$ background images, $50$ sternum images, $50$ rib images, $50$ transition region images, and $500$ cartilage images.

\begin{equation} \label{eq_classification_metrics}
\begin{split}
&Accuracy = \frac{TN+TP}{TN+TP+FN+FP} \\
&Sensitivity = \frac{TP}{TP+FN} \\
&Specificity = \frac{TN}{TN+FP} \\
\end{split}
\end{equation}

\begin{table}[htb!]
    \begin{center}
    \caption{Results of bone classification}%
    \label{tab:classification}
    \resizebox{0.48\textwidth}{!}{
    \begin{tabular}{ccccc}%
    \noalign{\hrule height 1.2 pt}
         \bfseries{ Class } & \bfseries{Accuracy} & \bfseries{Sensitivity} & \bfseries{Specificity} &\bfseries{AUC}\\
         \hline
         Background & 1.0 & 1.0 & 1.0 & 1.0  \\
         Sternum & 0.96  & 0.48 & 1.0 & 0.96\\
         Rib & 0.95 & 0.35 & 1.0 & 0.96 \\
         Transition & 0.87 & 0.67 & 0.89 & 0.89\\
         Cartilage & 0.96 & 0.97 & 0.93 & 0.98 \\
    \noalign{\hrule height 1.2 pt}
    \end{tabular}
    }
    \end{center}
\end{table}

\par
It can be seen from Table~\ref{tab:classification} that high numbers are reported for cartilage images in terms of different metrics. This means that the classification network can properly identify the cartilage from others. The classification results of background images achieved the best performance compared with other classes. The AUC results computed on the confusion matrix for all five classes also indicate the well-trained model can properly predict the classes. Among the results, the classification results of the transition part are relatively poorer than others. This is because connection parts only have ambiguous boundaries and are prone to have mixed characteristics from the two connected classes, which leads to more false results. For $100$ transition region images, $67$ images are successfully identified, while $28$ and $5$ images are wrongly classified as cartilage and rib bones, respectively. Although the classification accuracy of the sternum and ribs reach $0.96$ and $0.95$, respectively, the sensitivity is only $0.48$ and $0.35$ in this study. This is because true positive identification of sternum and ribs times is relatively low. In both cases, a large partial of the images is wrongly classified into transition class ($52$ and $65$ of sternum and ribs, respectively). Regarding the cartilage bone, the computed sensitivity and AUC are $0.97$ and $0.98$, respectively, on $1000$ images from two unseen volunteers. The good performance of cartilage bone classification can further boost the fine segmentation performance. Moreover, good classification results of cartilage images are the base for creating a high-quality US point cloud of patients for further registration to transfer the planned paths.

\subsection{Bone Segmentation Performance}
\par
To investigate the potential improvement caused by the explicit consideration of classification results, we first compared the segmentation results obtained by the coarse and fine segmentation networks on $1400$ images from two unseen patients. The results are depicted in Table~\ref{tab:segmentation_results}. Regarding coarse segmentation, we can find that the performance using DeepLabV3+ is better than a classical U-Net in terms of both the Dice coefficient ($0.69$ vs $0.53$) and IoU ($0.61$ vs $0.43$) on unseen data. To validate the improvement after further incorporating the classification information using CAM, the results predicted by the fine segmentation network on the same dataset ($1400$ mixed images) are computed. A slightly improvement is witnessed from Table~\ref{tab:segmentation_results}. The Dice coefficient and IoU are enhanced to $0.72$ and $0.64$, respectively.

\par
Since the precise segmentation of cartilage bone plays a key role in following registration tasks, we further trained a separate fine segmentation model only for extracting the cartilage bone. During the inference period, this fine cartilage segmentation model will only be triggered when the input images are identified as cartilage bone. The segmentation results on $1000$ unseen cartilage images reach $0.88$ and $0.79$ in terms of the Dice coefficient and IoU, respectively. 
An intuitive illustration of the autonomous segmented cartilage bone of an unseen patient can be found in 3D in Figs.~\ref{fig_coarse_alignment} and \ref{fig_fine_alignment}.

\begin{table}[htb!]
    \centering
    \caption{The summary of the segmentation performance (mean$\pm$SD)}%
    \label{tab:segmentation_results}
    \begin{tabular}{ccc}%
    \noalign{\hrule height 1.2 pt}
         \bfseries{Segmentation} & \bfseries{Dice} & \bfseries{IoU}  \\
         \hline
         Coarse: U-Net & 0.53 $\pm$ 0.34 & 0.43 $\pm$ 0.32  \\
         Coarse: DeepLabV3+ & 0.69 $\pm$ 0.34 & 0.61 $\pm$ 0.32  \\
         Fine (mixed classes) & 0.72 $\pm$ 0.32 & 0.64 $\pm$ 0.30 \\
         \hline
         Fine cartilage: & 0.88 $\pm$ 0.09 & 0.79 $\pm$ 0.13\\
    \noalign{\hrule height 1.2 pt}
    \end{tabular}
\end{table}


\subsection{VAE-based Geometry-Aware Post-Processing Performance}
\par
\begin{figure}[ht!]
\centering
\includegraphics[width=0.40\textwidth]{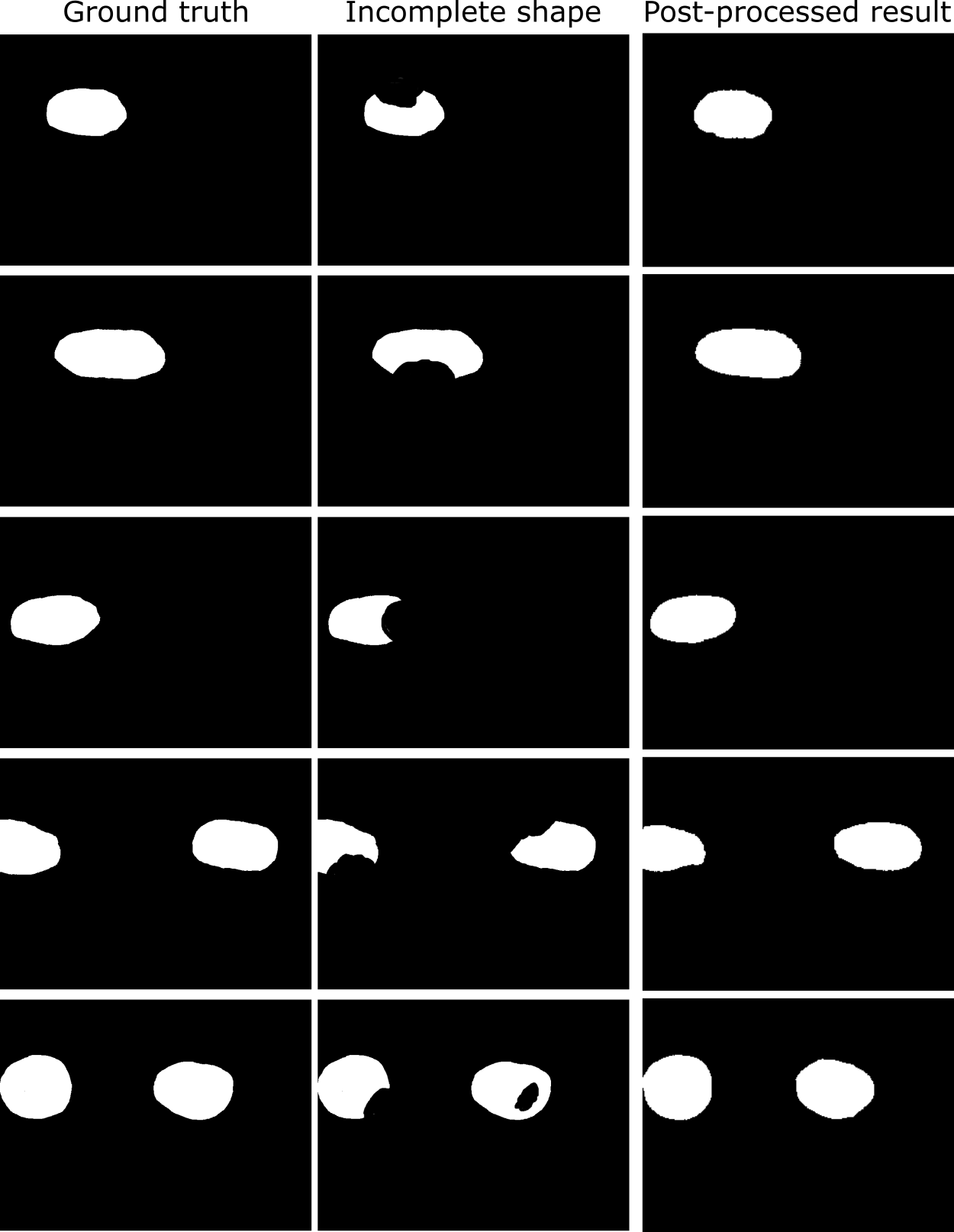}
\caption{The illustration of the VAE-based boundary-constraint postprocessing results in various cases.
}
\label{Fig_lVAE_postprocess}
\end{figure}

To ensure the obtaining of precise cartilage bone geometry, a VAE-based postprocessing approach is applied to the predicted binary mask of fine segmentation results. To intuitively demonstrate the effectiveness of the postprocessing approach, a few representative results are shown in Fig.~\ref{Fig_lVAE_postprocess}. The incomplete shape masks $M_{is}$ were manually decayed from the ground truth masks $M_{gt}$ of unseen cartilage bone images. Then, following the procedures described in Sec.~\ref{sec:VAE_bounday}, the incomplete shape mask is fed to the VAE encoder to compute the feature vector $Z$ in latent space. Then, the nearest sample $Z^{\prime}$ among the augmented validated samples ($110K$) is used as the approximation of feature representation for the VAE decoder. It can be seen from Fig.~\ref{Fig_lVAE_postprocess} that the processed cartilage geometry masks $M_{pr}$ are significantly improved when the decay happens in different parts of the anatomy of interest.

\par
To quantitatively evaluate the performance, the Dice coefficient ($1-L_{Dice}$) was computed twice between the $M_{gt}$ and $M_{is}$, and $M_{gt}$ and $M_{pr}$, respectively. For the five representative examples in Fig.~\ref{Fig_lVAE_postprocess}, the computed Dice coefficients are $0.87$ vs $0.90$, $0.88$ vs $0.95$, $0.82$ vs $0.93$, $0.88$ vs $0.94$ and $0.93$ vs $0.96$, respectively. The results demonstrated the presented VAE-based postprocessing can help to further guarantee the geometry of cartilage bone segmentation. 


\subsection{Intercostal Scanning Path Transferring Performance}
\par
\begin{figure*}[ht!]
\centering
\includegraphics[width=0.90\textwidth]{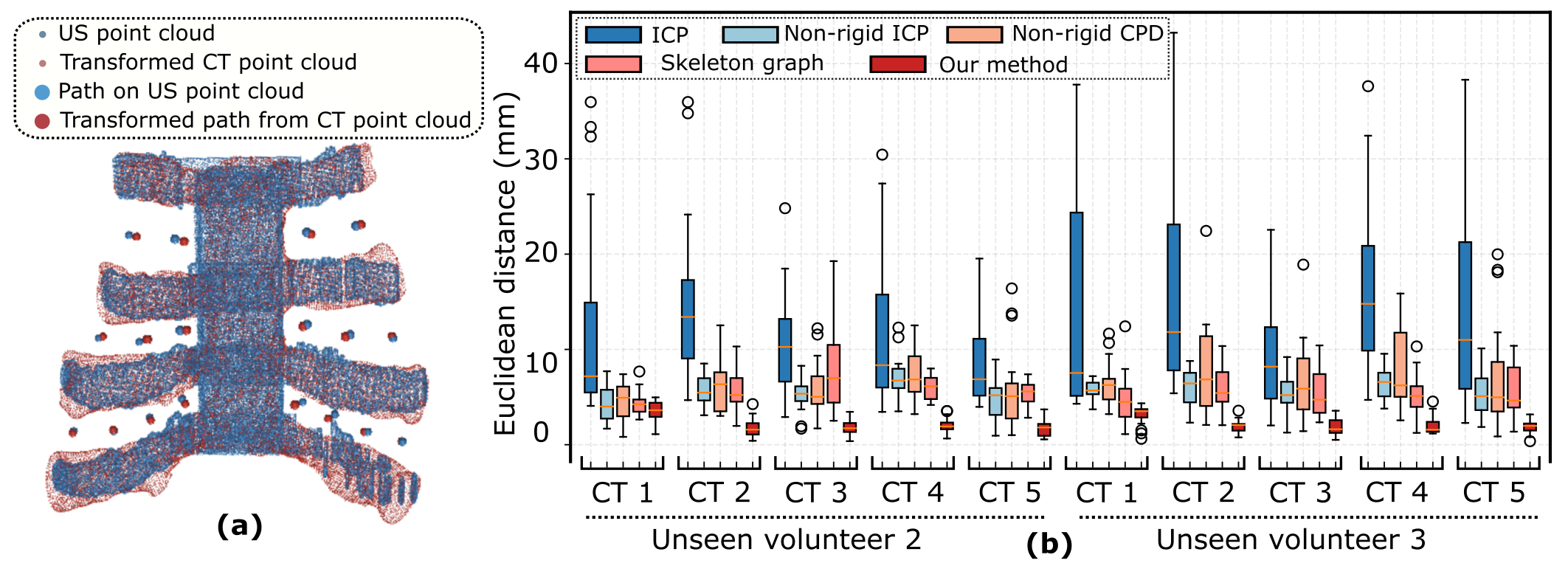}
\caption{Performance of intercostal paths transferring from CT to US space. (a) An representative results illustrating the mapped 18 intercostal waypoints from CT to US space. (b) The statistical path transferring results, in terms of Euclidean distance, computed using the proposed method and other existing approaches based on five CTs and two unseen volunteers' thoracic images.
}
\label{Fig_path_transfer}
\end{figure*}

To quantitatively validate the effectiveness of the whole system on intercostal path transferring, five CT chest volumes from a public dataset and two autonomously extracted US volumes from unseen volunteers were used. The same protocol was used to determine $18$ waypoints on each cartilage point cloud from CT and US. According to the length of individual cartilage (2nd, 3rd, 4th, and 5th), $2$, $3$, and $4$ waypoints were generated for the three intercostal spaces at each side [see Fig.~\ref{Fig_path_transfer}~(a)]. For the utmost assurance of having matched waypoints from CT and US point clouds, cartilage bone was carefully annotated for this validation. Then, KNN is used to extract predefined number clusters and their centroid point. By connecting the corresponding centroids in neighboring cartilage, the midpoints of the connecting line are adopted as waypoints in the intercostal space.

\par
To quantitatively evaluate the performance of intercostal path mapping from CT to US, the Euclidean error $E_{euc}$ is computed between the transformed waypoints from CT and the ones defined on US point cloud in this study. In order to investigate the impact of the learning-based bone segmentation on the final registration performance, $E_{euc}$ is computed for each CT using two US point clouds (manually annotated and autonomously segmented) obtained from the same volunteer. The results [mean (SD)] have been depicted in Table~\ref{tab:path_transfer}. In most cases (CT 1, 2, 4, and 5), the results obtained using the manually labeled cartilage are slightly better than the ones obtained using the autonomous segmentation approach. The average difference over all five CTs is only $0.5~mm$. 

\par
To further validate the overall path mapping performance on unseen volunteers, the proposed non-rigid skeleton graph-based registration was repeatedly used to register the autonomously segmented US point clouds from volunteers 2 and 3 to different CTs. The results in Table~\ref{tab:path_transfer} show that the similar $E_{euc}$ is obtained for individual CT. In particular, for CTs 2 and 5, the best mapping performance is achieved for unseen volunteer 2. This demonstrates the proposed bone segmentation and post-process network are sufficient to be used for efficiently mapping the preplanned path from CT to US space.

\begin{table}[htb!]
    \begin{center}
    \caption{The performance of intercostal path transferring in terms of Euclidean distance  [Mean(SD)]}%
    \label{tab:path_transfer}
    \resizebox{0.48\textwidth}{!}{
    \begin{tabular}{cccccc}%
    \noalign{\hrule height 1.2 pt}
         \bfseries{ Subjects } & \bfseries{CT1} & \bfseries{CT2} & \bfseries{CT3} & \bfseries{CT4} & \bfseries{CT5}\\
         \hline
         Volunteer 1~$\bigstar$ & $\textbf{2.2}$ ($0.8$) & $2.0$ ($0.8$) & $1.9$ ($0.7$) & $2.0$ ($0.8$) & $1.9$ ($0.9$) \\
         Volunteer 1 & $3.6$ ($1.5$) & $2.2$ ($1.1$) & $\textbf{1.7}$ ($0.9$) & $2.7$ ($0.9$) & $2.1$ ($1.1$) \\
         Volunteer 2 & $3.4$ ($1.3$) & $\textbf{1.8}$ ($1.0$) & $1.8$ ($0.8$) & $2.0$ ($0.8$) & $\textbf{1.8}$ ($0.9$) \\
         Volunteer 3 & $3.1$ ($1.0$) & $1.9$ ($0.7$) & $1.9$ ($0.8$) & $2.0$ ($0.9$) & $1.9$ ($0.7$) \\
    \noalign{\hrule height 1.2 pt}
    \multicolumn{6}{l}{Unit: mm; ~~$\bigstar$ indicates the manual bone annotation}  
    \end{tabular}
    }
    \end{center}
\end{table}


\par
To further investigate the performance of the proposed non-rigid dense skeleton graph-based registration method, the classic ICP\cite{besl1992method}, non-rigid ICP~\cite{amberg2007optimal}, non-rigid CPD~\cite{myronenko2010point} and the keypoint-based skeleton graph method were tested as well. For the latter three nonrigid approaches, the mapping of the waypoints from CT to US space was carried out in the same way as this study. A sphere region (radius is $20~mm$) around individual waypoints is used to compute the local transformation matrix. The results on two unseen volunteers are summarized in Fig.~\ref{Fig_path_transfer}.

\par
It can be seen from Fig.~\ref{Fig_path_transfer} that the presented dense skeleton graph-based registration approach can outperform its peers in the scenarios of thoracic application. The mapping errors computed using the presented methods ($2.2\pm1.1~mm$) are significantly smaller than the ones obtained using other methods for different CTs and volunteers' US data (ICP: $13.2\pm9.6~mm$, Non-rigid ICP: $5.6\pm2.0~mm$, Non-rigid CPD: $6.6\pm3.9~mm$, and Keypoint-based skeleton graph $5.6\pm2.5~mm$). The second-best results obtained by the keypoint-based skeleton method are two times larger than the ones obtained by the presented dense graph-based method. The results obtained by the classic ICP are the worst across all cases, and the errors are far larger than others. This is because the classical ICP is more sensitive to the difference between source and target point clouds. A few outliers will significantly impair the overall ICP results. 
Similar findings can also be witnessed in other methods (non-rigid ICP, CPD, and keypoint-based skeleton graph), while the one using the dense graph-based registration method is the least affected in our setup, thanks to the use of a dense graph. A representative intercostal path mapping results computed between the CT1 and volunteer 2 are depicted in Fig~\ref{Fig_path_transfer}~(a). In addition, the results computed based on two unseen volunteers' data are consistent, which demonstrates that proposed segmentation and registration have the potential to adapt inter-patient variations.  

\par
In addition, we computed the time efficiency for each part. The average inference time for the coarse segmentation module and classification are $27~ms$ and $9~ms$, respectively, across $1400$ images. The classification and fine segmentation together require $149~ms$ in total, with CAM generation taking $90~ms$ and fine segmentation is $59~ms$ on average for individual images. The boundary-constrained VAE process is only applied for cartilage images, with a computational time averaging $756~ms$ across $1000$ images.





\section{Discussion}
\par
This work presents a pipeline for mapping a pre-planned scanning path from CT/MRI template onto individual patients, facilitating autonomous robotic US examination. We showcase its effectiveness through the challenging intercostal examination, where a limited acoustic window is encountered. It is worth noting that this method holds promise beyond intercostal application; it can autonomously generate scanning paths for US examination of other abdominal applications by mapping the rib skeleton from CT to US. In addition, the presented VAE-based boundary-constrained method can be extended for shape completion in various applications. The current study only contains a few waypoints in the intercostal space. In real scenarios, a continuous scanning path can be generated using advanced RL framework~\cite{bi2024autonomous}, which can be directly used for further robot-assisted US image scanning. Furthermore, it's worth highlighting that the CT template need not be singular. A comprehensive template library could include templates from individuals of different genders, ages, BMI, heights, ethnicities, and so forth. To address practical challenges, existing studies that address potential patient movement~\cite{jiang2021motion, jiang2022precise} and force-induced deformation~\cite{jiang2021deformation, jiang2023defcor} during scanning can be further integrated to develop a fully autonomous RUSS.








\section{Conclusion}
\par
This study presents a method to autonomously and precisely map the scanning path from a tomographic template to individual patients. It leverages a class-aware cartilage US bone segmentation network and a non-rigid skeleton graph-based registration method that takes into account the subcutaneous bone structure. To achieve accurate and plausible geometry of the cartilage bone, the CUS-Net consists of four modules: coarse segmentation, classification, fine segmentation, and geometry-constraint VAE-based post-processing. Based on the results of two unseen volunteers' thoracic images, the final cartilage bone segmentation is improved from $0.69\pm0.34$ to $0.88\pm 0.09$ in terms of Dice and from $0.61\pm0.32$ to $0.79\pm0.13$ in terms of IoU. In addition, we can find a significant enhancement in the geometry completeness and plausibility after applying the VAE-based post-processing. The method's efficacy was further validated through joint validation on five CT templates, where patient-specific cartilage point clouds were extracted from two unseen volunteers. The results demonstrate the proposed method is more precise and robust than other approaches in all ten combination cases. The results demonstrate that the proposed method can outperform the classical ICP, non-rigid ICP, and CPD and Keypoint-based skeleton graph algorithms in our setup in terms of Euclidean distance for path transferring error ($2.2\pm1.1~mm$ vs. $13.2\pm9.6~mm$, $5.6\pm2.0~mm$, $6.6\pm3.9~mm$ and $5.6\pm2.5~mm$). These results affirm the feasibility of autonomously and accurately mapping the scanning path for challenging thoracic applications, such as intercostal liver examination, using the proposed approach. Future studies will expand on this method by testing it on various thoracic applications, incorporating specific anatomy information to enhance registration performance, and integrating multi-modal registration techniques to further optimize the transferred scanning path.



\bibliographystyle{IEEEtran}
\bibliography{IEEEabrv,references}

\end{document}